# Spectral Analysis of Marine Debris in Simulated and Observed Sentinel-2/MSI Images using Unsupervised Classification

Bianca Matos de Barros[1], Douglas Galimberti Barbosa[2], Cristiano Lima Hackmann[3]

B.M.B: ORCID® 0000-0003-0818-1267
D.G.B.: ORCID® 0000-0003-2257-3195
C.L.H.: ORCID® 0000-0001-9225-5980

[1,2,3] Universidade Federal do Rio Grande do Sul. Programa de Pós-Graduação em Sensoriamento Remoto. Av. Bento Gonçalves, 9500, CEP 91501-970. Porto Alegre.

* Corresponding author: bianca.matos@ufrgs.br

**ABSTRACT**

Marine litter poses significant threats to marine and coastal environments, with its impacts ever-growing. Remote sensing provides an advantageous supplement to traditional mitigation techniques, such as local cleaning operations and trawl net surveys, due to its capabilities for extensive coverage and frequent observation. In this study, we used Radiative Transfer Model (RTM) simulated data and data from the Multispectral Instrument (MSI) of the Sentinel-2 mission in combination with machine learning algorithms. Our aim was to study the spectral behavior of marine plastic pollution and evaluate the applicability of RTMs within this research area. The results from the exploratory analysis and unsupervised classification using the KMeans algorithm indicate that the spectral behavior of pollutants is influenced by factors such as the type of polymer and pixel coverage percentage. The findings also reveal spectral characteristics and trends of association and differentiation among elements. The applied methodology is strongly dependent on the data, and if reapplied in new, more diverse, and detailed datasets, it can potentially generate even better results. These insights can guide future research in remote sensing applications for detecting marine plastic pollution.

**KEYWORDS**: MARINE DEBRIS. REMOTE SENSING. RADIATIVE TRANSFER MODELS. IMAGE CLASSIFICATION. CLUSTERING ALGORITHM.

## INTRODUCTION

Marine and coastal ecosystems are under significant threat from factors such as warming, sea level rise, and pollution. A notable element of the pollution issue is the continuous influx of slowly degrading materials into the oceans, which over time leads to their gradual accumulation. It is estimated that plastic materials make up approximately 60% to 90% of all marine debris (Baker et al. 2016, Maximenko et al. 2019).

Remote sensing techniques can complement traditional approaches, such as cleanup operations and trawl surveys, by providing extensive coverage and frequent observation. However, these studies often encounter limitations, particularly concerning the resolution of sensors and the availability of remotely acquired images validated in situ. Such images are crucial for the training and validation of image classification models.

Current strategies to overcome these challenges include using images without in situ validation (Biermann et al. 2020), testing radiometric indexes (Themistocleous et al. 2020), and combining datasets from multiple sensors simultaneously (Topouzelis et al. 2019, Topouzelis et al. 2020, Papageorgiou et al. 2022). These are valid approaches, but they often introduce further uncertainties or limitations into the results.

An alternative is the use of Radiative Transfer Models (RTMs) to create simulated scenes with different configurations and scenarios. This approach is common in remote sensing, especially in the context of vegetation and agricultural applications (Jacquemoud et al. 2000, Duthoit et al. 2008, Ali et al. 2021).

In this study, we combine remotely acquired observational data and RTM simulated data to conduct an exploratory analysis. We apply unsupervised machine learning algorithms to assess patterns of association and the impacts of using both types of data. Our main objective is to evaluate the potential and limitations of using remote sensing images and methods to detect marine plastics. To achieve this, we have two specific goals: i) to study the spectral behavior of marine pollution; ii) to assess the applicability of RTMs in the remote detection of plastic debris.

## METHODS

This section presents the methodology. It comprises three stages:

i. Data, where we select and process the observed database, and also create the synthetic scenes.

ii. Preprocessing, where we add radiometric indices as new features to the datasets, clean the data and perform exploratory analysis and feature selection.

iii. Classification, where we perform unsupervised classification with the K-Means algorithm and evaluate its results.

### DATA

#### Study area

The study area of the observed images was the Greek island of Lesbos, located in the northeastern part of the Aegean Sea (Figure 1).

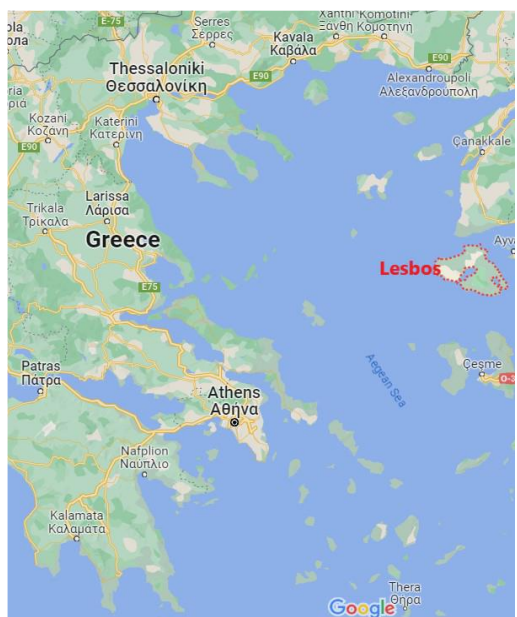

**Figure 1.** Map of the study area on the Greek island of Lesbos (highlighted in red).
Source: Adapted from Google Maps.

### *Observed remote sensing data selection*

We sought images acquired by orbital remote sensing that had in situ confirmation of plastic presence in coastal regions. The selection criteria were the in situ confirmation and the availability of target georeferencing data, to avoid uncertainties or limitations in classification later. We selected all images that matched these criteria. The selected datasets are described in the following paragraphs. All images were acquired using the Sentinel-2/MSI sensor.

*Plastic Litter Project 2019.* The Plastic Litter Project (PLP) has carried out annual experiments to implant artificial plastic targets during the overpass of satellite over Greek beaches since 2018. The Marine Remote Sensing Group (MRSG), affiliated with the Department of Marine Sciences at the University of the Aegean, is responsible for the project. In the 2019 edition, a set of artificial targets was built, in sizes of 1 x 5 meters and 5 x 5 meters, with modular connections, allowing their combination in different formats, composition materials, and coverage percentages, with remote data acquisition in different configurations and on different dates throughout 2019 (Figure 2, top). Target materials included transparent bottles (Polyethylene Terephthalate, PET), plastic bags (Low-Density Polyethylene, LDPE), and natural debris (reeds, Arundo donax).

The targets were deployed in the waters near Tsamakia beach in Mytilene, on the Greek island of Lesvos, and georeferenced using Global Positioning System (GPS) devices to obtain their exact location. The MSI sensor on board the Sentinel-2 satellite and a camera with a higher geospatial resolution, mounted on a Unmanned Aerial Vehicle (UAV), acquired images simultaneously. The calculation of percentages of plastic coverage in each pixel of satellite images was based on photographs acquired by UAV. The authors provided the MSI/Sentinel-2 data, the geolocation of the targets, and the percentages of plastic presence for each pixel. This data can be obtained from the MRSG website or the Zenodo database (Topouzelis 2020). This allows researchers to locate the experiment images, map the pixels containing the plastic targets, extract the spectral information, and relate them to the respective coverage percentages of each class (Topouzelis et al. 2019). The 2019 selected dates were 2019/04/18, 2019/05/03, 2019/05/18, 2019/05/28, and 2019/06/07, which made up the entire dataset.

*Plastic Litter Project 2021.* In the 2021 edition of the PLP, targets were deployed near Skala Loutron, also on the Greek island of Lesvos. Two large targets measuring 28 meters in diameter, one covered in a white mesh of High Density Poly Ethylene (HDPE) and the other covered in planks, were anchored on the beach for four months (Figure 2, bottom), allowing multiple acquisitions by Sentinel-2.

The authors created orthophotographic maps of the implantation area using UAV images and published them in the Zenodo (Papageorgiou & Topouzelis 2022) database. The geolocation of the targets in the images can be confirmed using these orthophotographic maps. Effects of submersion and bioencrustation over time are also observable. However, the data provided by the authors did not include the percentages of plastic coverage in each pixel of the image (Topouzelis et al. 2020). The 2021 selected dates were 2021/06/21, 2021/07/01, 2021/07/06, 2021/07/21, and 2021/08/25. These are dates whose images were representative of the set, which were in good weather conditions and demonstrated the transformations in the targets over time. The variation among the dates includes different degrees of targets submergence and biofouling.

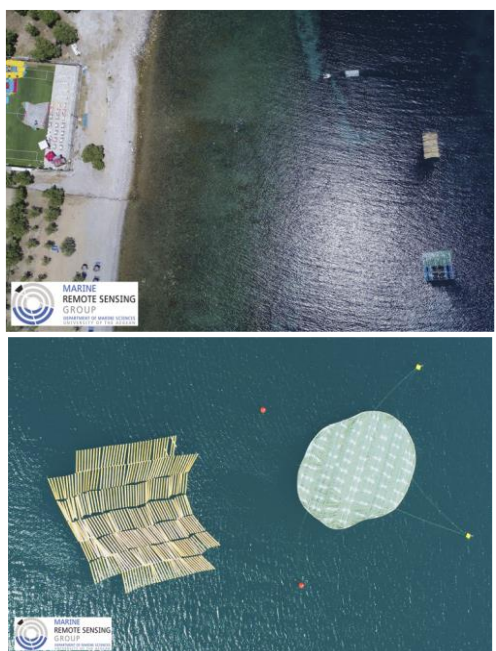

**Figure 2.** Artificial targets deployed in Greece during PLPs 2019 (top) and 2021 (bottom) in aerial photographs acquired by UAV. Sources: https://tinyurl.com/4yyr6s88 (top); https://tinyurl.com/ypmtsajb (bottom).

### Observed data processing

The MSI/Sentinel-2 level L1 products for the selected images were obtained from the United States Geological Survey portal[1] (USGS) and from the Copernicus Open Access Hub[2]. The Acolite processor[3] was used for atmospheric correction, applying the DSF algorithm to the L1 level images. Acolite also resampled all image bands to 10 meters, the best available spatial resolution.

The Sentinel Application Platform (SNAP) software software[4], which is a common platform designed for processing data products from Sentinel missions, was used for cropping the regions of interest in each image.

---

[1] Available at https://earthexplorer.usgs.gov/

[2] Available at: https://scihub.copernicus.eu/dhus/

[3] Available at: https://github.com/acolite/acolite/releases/tag/20220222.0

[4] Available at https://step.esa.int/main/download/snap-download/

From each date, two subsets were extracted: one containing the targets and the surrounding water, usually in 12x12 pixels (a little bigger on some dates), and another containing a stretch of the coastal land surface, with a size of 10x10 pixels. Subsets were exported to TIFF format.

For the 2019 dates, the geolocation of the plastic targets and its percentual coverage for each pixel were based on the data available for download (Topouzelis et al. 2019). For the 2021 dates, the geolocation of the targets used the georeferenced orthophotos, also available for download (Topouzelis et al. 2020). Its percentual coverage for each pixel was estimated using visual inspection in two categories: 100% (full pixel coverage) or less than 100% (partial coverage, unknown percentage). Geolocation data and the respective TIFF subsets were superimposed using The QGIS software[5] for pixel mapping. A script coded in Python integrated spatial and spectral data. Only pixels with 100% water or sand coverage were labeled with these classes. Pixels containing any percentage of plastic were labeled as Plastic.

### Simulated remote sensing data creation

*Polymers selection*. The creation of simulated images requires a set of polymers with their respective spectral signatures. In this work, we selected a set of polymers from (Garaba & Dierssen 2018): the virgin pellets of Polyamide 6 and 6.6 or Nylon (PA 6 and PA 6.6), Polyvinyl chloride (PVC), Low-density polyethylene (LDPE), PET, and Polypropylene (PP), and also a mean signature made of several microplastics collected in the Pacific Ocean (μ-NAPO). The spectral databases were published and are available online (Garaba and Dierssen 2019a; Garaba and Dierssen 2019b; Garaba and Dierssen 2017). We chose these polymers as samples of the most frequently found plastics in coastal and marine environments. Their signatures can be seen in Figure 3. The signatures of selected elements have unique characteristics, with absorption zones centered at 931, 1215, 1417, and 1732 nm.

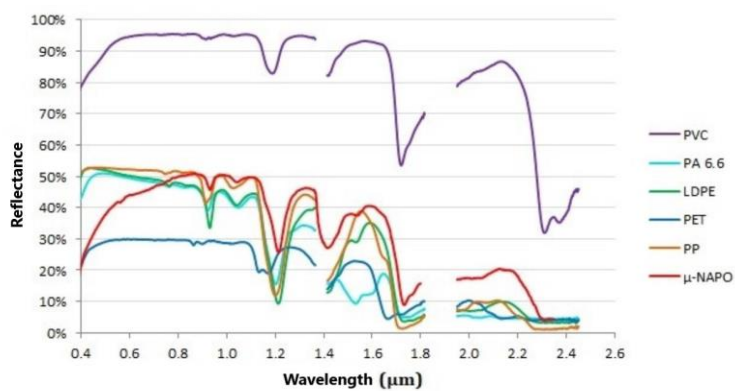

**Figure 3.** Spectral signatures of μ-NAPO (mean signature of microplastics collected in the Pacific Ocean) and of the virgin polymers PVC, PA 6.6, LDPE, PET and PP. Source: Adapted from Garaba & Dierssen, 2019a and Garaba & Dierssen, 2017.

*Creation of simulated scenes.* The simulated dataset was generated using the DART (Discrete Anisotropic Radiative Transfer) computational tool (Gastellu-Etchegorry et al. 2012). We constructed threedimensional scenes each covering an area of 1200 m x 1200 m, totaling 720 km2. The simulated sensor was the MSI/Sentinel-2, featuring a 12-bit radiometric resolution, a revisit frequency of 5 days, and a spatial resolution varying from 10 to 60 m (Table 1). The sensor bands cover the Visible (VIS), Red Edge (RE), Near Infrared (NIR) and Shortwave Infrared (SWIR) regions of the electromagnetic spectrum. For consistency in nomenclature between simulated and observed data,

---

bands B8 and B8A, referred to as NIR and RedEdge4 in Table 1, are labeled as NIR1 and NIR2, respectively, in this study.

Simulated square plastic objects were placed on the surface of water bodies, with areas of $100m^2$ (covering 100% of the 10m MSI sensor pixel), $64m^2$ (80%), $16m^2$ (40%), $36m^2$ (60%), $4m^2$ (20%). These objects incorporated the plastic spectral signatures described in the Subsection 'Polymer Selection', with water and sand spectral signatures provided by DART.

We only considered bands with spatial resolutions of 10 and 20 m in this study (Blue, Green, Red, Red-Edge1, RedEdge2, RedEdge3, NIR1, NIR2, SWIR1, SWIR2). A total of 30 synthetic images were created, one for each of the five coverage ratios for each of the six selected plastic spectral signatures. DART generated a product for each of the ten bands, leading to 300 simulated remote sensing products.

**Table 1.** MSI/Sentinel-2 bands.

| Band – Name | Region | Central wavelength | Resolution |
|---|---|---|---|
| B1 – Aerosols | – | 443.9 – 442.3 nm | 60 m |
| B2 – Blue | VIS | 496.6 – 492.1 nm | 10 m |
| B3 – Green | VIS | 560 – 559 nm | 10 m |
| B4 – Red | VIS | 664.5 – 665 nm | 10 m |
| B5 – Red Edge 1 | RE | 703.9 – 703.8 nm | 20 m |
| B6 – Red Edge 2 | RE | 740.2 – 739.1 nm | 20 m |
| B7 – Red Edge 3 | RE | 782.5 – 779.7 nm | 20 m |
| B8 – NIR | NIR | 835.1 – 833 nm | 10 m |
| B8A – Red Edge 4 | NIR | 864.8 – 864 nm | 20 m |
| B9 – Water vapor | – | 945 – 943.2 nm | 60 m |
| B10 – Cirrus | – | 1373.5 – 1376.9 nm | 60 m |
| B11 – SWIR 1 | SWIR | 1613.7 – 1610.4 nm | 20 m |
| B12 – SWIR 2 | SWIR | 2202.4 – 2185.7 nm | 20 m |

Source: Adapted from https://tinyurl.com/u8evfn7k.

Each scene of bands with a spatial resolution of 20m contains 60 pixels x 60 pixels, totaling 3,600 pixels (256 plastic, 1,672 water, and 1,672 sand). For bands with a spatial resolution of 10 m, each scene contains 120 pixels x 120 pixels, totaling 14,400 pixels (256 plastic, 7,072 water, and 7,072 sand). The size and position of plastic objects were consistent across both spatial resolutions.

We incorporated the spatial and spectral information of the simulations into a script coded in Python. To compile pixel information in a database, we had to use a single spatial resolution, so we resampled the pixels of the 20 m bands to the 10 m resolution. We tested the nearest neighbor, the bilinear interpolation and the cubic interpolation methods[6]. The bilinear interpolation method was selected due to its smaller amplitude compared to the other methods, resulting in a closer match to the distribution of the observed data. Additionally, simulated dataset values were rounded to four decimal places.

Only pixels with 100% water or sand coverage were labeled with these classes. Pixels containing any percentage of plastic were labeled as Plastic.

---

[6] We used using the class *Resampling*, documentation available at: https://rasterio.readthedocs.io/en/latest/topics/resampling.html

## PREPROCESSING

### *Radiometric Indices*

A set of radiometric indices from previous studies were calculated for both simulated and observed data to assess their applicability. These indices included the Normalized Difference Water Index (NDWI, Equation 1), Water Ratio Index (WRI, Equation 2), Normalized Difference Vegetation Index (NDVI, Equation 3), Automated Water Extraction Index (AWEI, Equation 4), Modified Normalization Difference Water Index (MNDWI, Equation 5), Simple Ratio (SR, Equation 6), Plastic Index (PI, Equation 7), Reversed Normalized Difference Vegetation Index (RNDVI, Equation 8) (Themistocleous et al. 2020) and Floating Debris Index (FDI, Equation 9) (Biermann et al. 2020).These indices were incorporated as new features in both datasets.

$$\text{NDWI} = \frac{B3 - B8}{B3 + B8} \tag{1}$$

$$\text{WRI} = \frac{B3 + B4}{B8 + B12} \tag{2}$$

$$\text{NDVI} = \frac{B8 - B4}{B8 + B4} \tag{3}$$

$$\text{AWEI} = 4(B3 - B12) - 0.25B8 - 2.75B11 \tag{4}$$

$$\text{MNDWI} = \frac{B3 - B12}{B4 + B12} \tag{5}$$

$$\text{SR} = \frac{B8}{B4} \tag{6}$$

$$\text{PI} = \frac{B8}{B8 + B4} \tag{7}$$

$$\text{RNDVI} = \frac{B4 - B8}{B4 + B8} \tag{8}$$

$$\text{FDI} = R_{rs,NIR} - R'_{rs,NIR} \tag{9}$$

where

$$R'_{rs,NIR} = R_{rs,RE2} + \alpha \cdot \beta \cdot 10$$

$$\alpha = R_{rs,SWIR1} - R_{rs,RE2}$$

$$\beta = \frac{(\lambda_{NIR} - \lambda_{RED})}{(\lambda_{SWIR1} - \lambda_{RED})}$$

### Data cleaning

The datasets presented no noticeable inconsistencies, incompleteness, noise, redundancies, or absences. However, division by zero instances occurred when the indices were added, affecting 512 water pixels in the simulated dataset and 20 pixels in the observed dataset (one plastic and 19 water). These quantities were relatively small compared to the total dataset, thus the affected samples were disregarded in the subsequent steps.

### Exploratory analysis

Exploratory data analysis was fundamentally based on a set of visual methods capable of describing the datasets. We analyzed measures of location, spread, and distribution for both datasets and examined the relationships between them.

The selection of graphics was guided by the following questions:

- What are the distributions of classes within both datasets?
- How do the observed and simulated datasets compare and contrast?
- Which features or characteristics most significantly account for the similarities and differences between the observed and simulated datasets?

The Kolmogorov-Smirnov test (Dodge 2008) was applied to both the observed and simulated datasets to assess whether the samples from the two datasets originate from the same distribution.

### Features selection

Feature selection utilized the Random Forest (RF) classifier[7], an estimator that generates rankings by combining data sampling techniques with Decision Trees. Decision Trees are models that predict target variables by learning simple decision rules inferred from input features. RF was chosen because, in addition to classification, it also generates an importance score for each feature, indicating its influence on the result: the higher the score, the more relevant the feature. The algorithm was applied twice, using the following datasets:

- Full set of simulated data;
- Set containing only the water and plastic classes from the simulated data.

The datasets were split into 75% for training and 25% for testing. The maximum depth value selected for the trees was 3 levels, as it is the smallest value to reach maximum accuracy in the classification of simulated data. The training and testing process was the same for both datasets. In the end, the feature importance calculations for each of the datasets were graphed.

Feature importances were compared with their correlation matrices. The correlation was analyzed both for the entire dataset and on a class-by-class basis, aiming to select sets of features with high scores assigned by the RF and low correlation among them.

## CLASSIFICATION

### Unsupervised classification

The unsupervised classification algorithm used was KMeans, which clusters the data by attempting to separate the samples into n groups of equal variance. It was implemented using Scikit-learn. K-Means divides a set of samples into disjoint clusters, with each cluster described by the mean of the samples it contains. These means are commonly referred to as "centroids." K-Means aims to choose centroids that minimize the inertia, or the sum of squared distances within each cluster (Equation 10). The algorithm requires the number of clusters to be specified.

---

[7] Documentation available at: https://scikit-learn.org/stable/modules/generated/sklearn.ensemble.RandomForestClassifier.html

$$\sum_{i=0}^{n} \min_{\mu_j \in C} \left( \|x_i - \mu_j\|^2 \right) \qquad (10)$$

After initialization, K-Means consists of a two-step iteration: the first step assigns each sample to its nearest centroid, and the second step creates new centroids by taking the average value of all samples assigned to each previous centroid. The difference between the old and new centroids is calculated, and the algorithm repeats these two steps until this difference becomes less than a certain threshold. In other words, it continues until the centroids no longer move significantly.

The unsupervised classification with K-Means was tested using all sets of features and including all classes from the data, both for the simulated and observed sets. For both sets, the classification was initially performed with 3, 4, and 5 clusters, numbers close to the number of classes in the sets, in order to analyze the grouping trends. All classifications were performed with the same initialization to evaluate the variations caused by different initial configurations in both sets.

The cluster analysis was guided by the following questions:
- Did the number of clusters significantly affect the ranking for the same set of features?
- Which set of features classified the data with greater separability for each number of clusters?
- What are the characteristics of the pixels associated with the major and minor clusters? Are there any association patterns between different classes?

## RESULTS

### EXPLORATORY ANALYSIS

We applied the Kolmogorov-Smirnov test to the observed and simulated datasets, comparing feature by feature, both considering all classes simultaneously and in isolation. In all cases, the p-value was close to zero, indicating that, in all conditions tested, the observed and simulated datasets came from different distributions. This constitutes a violation of one of the basic assumptions of machine learning, which determines that the training, testing, and validation sets need to come from the same distribution. We assessed the performance of the classifiers considering that, in this scenario, the applicability of machine learning can be limited, considering that the function learned by the model is based on the statistical distribution of the training data and may not be generalizable to different distributions.

The simulated data first had 212,160 pixels of sand, 212, 160 pixels of water, and 7,680 pixels of plastic, totaling 432,000 samples. After pre-processing, there was a reduction in the number of water pixels to 211,648. The other classes maintained the original number of samples (Table 2). Among the plastic pixels, all polymers had the same number of samples (Figures 4 and 5).

**Table 2.** Description of simulated and observed datasets

| Class | Simulated Data (DART) | Observed Data (USGS) |
|-------|----------------------:|----------------------|
| Water | 211,648 | 1,943 |
| Sand | 212,160 | 0 |
| Coast | 0 | 1,069 |
| Plastic | 7,680 | 103 |
| Wood | 0 | 62 |
| **Total** | **431,488** | **3,177** |

The observed data originally had 3,197 samples, of which 3,177 were considered after pre-processing. As shown in Table 2, the observed dataset has 1,943 pixels of water (1,346 in 2019 and 597 in 2021), 1,069 pixels extracted from the Earth's surface coastal region (500 in 2019 and 569 in 2021) and 165 pixels of artificial targets suspended in the water, 103 of which are plastic (49 in 2019 and 54 in 2021) and 62 are wood (all from 2021). Among the plastic pixels, there was a predominance of those covered by HDPE mesh, from PLP 2021, in which all samples contained only this element. Among the others, from PLP 2019, there was a balanced distribution of pixels containing bags (LDPE) and bottles (PET) (Figure 5).

Both datasets are highly unbalanced. Class unbalances are also expected in natural scenes, mostly composed of surface cover and containing a sparse distribution of debris. The presence of different polymeric resins in the pixels representing plastic pollution is evenly distributed in the simulated data but unbalanced in the observed data. Considering that the data cquisition was artificially manipulated both in the simulated and observed data sets, greater variability and imbalance in the distribution of these elements in the environment can also be expected.

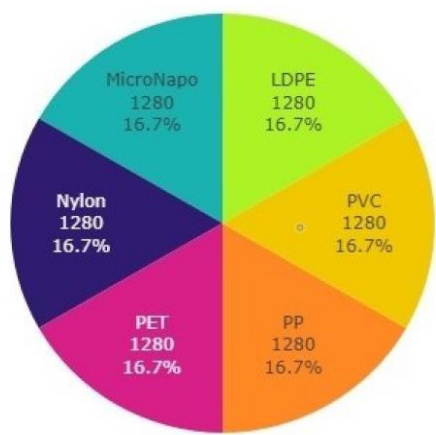

**Figure 4.** Distribution of polymers in plastic pixels in simulated dataset (DART).

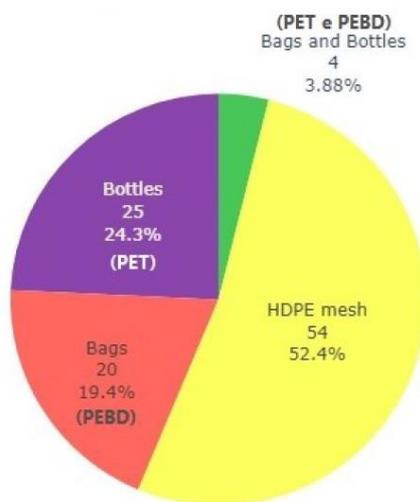

**Figure 5.** Distribution of polymers in plastic pixels in observed dataset (USGS).

The simulated data are highly correlated, as shown in Figure 6. Considering the total data set, with all classes, the correlation values approach 1 or -1 for all bands and almost all indices, WRI and FDI being the only exceptions.

The correlation between the bands in observed data is also high. Most of the bands correlate higher than 0.8 between them. The Blue band presents slightly lower correlations when compared to the other bands. On the other hand, the indices demonstrate smaller correlations both among themselves and in relation to the bands. In particular, the FDI, RNDVI, PI, SR, and NDVI indices have very low correlations with most other features. The lowest correlations between indices and bands are an indication of its potential and relevance for sample classification, supporting the methodology of the work.

All simulated scenes were in equal conditions as if their acquisition date were the same. The acquisition dates of the observed data were different, subjecting data to variations arising from climatic and environmental conditions and their respective effects on the final remote sensing products. The observed data are mostly composed of acquisitions made in 2019, with an overrepresentation of two dates: May 3 and 18, 2019 (Figure 7). On these dates, due to the different settings in the plastic targets, it was necessary to define a larger region of interest to capture the entire area.

The pixels of the coast and plastic classes are evenly divided between the two editions of the PLP. However, those of the wood class are only present in 2021. The pixels of the water class also have an overrepresentation of the year 2019, as a direct consequence of data collection in larger areas on May 3rd and 18th of that year.

The simulated data had a reflectance with an mean and a standard deviation higher than the observed data. When only the water and plastic classes were considered - as they were the only ones present in both sets - the mean of the simulated dataset was much closer to the mean of the observed dataset, although the standard deviation of the simulated dataset was still much higher than that of the observed dataset (Figure 8). The standard deviation of data acquired in 2021 was closer to the standard deviation of the simulated dataset than the standard deviation of data acquired in 2019.

Simulated dataset quartiles had a greater range than the observed dataset, with a big difference in variability between substrates (water and sand) than observed data (water and coast). In the observed data, the 2019 dataset shows more differences between the spectral behavior of the different classes than the 2021 data (Figure 9). In class-by-class comparison between simulated and observed data, the observed plastic had a much smaller variability than the simulated one. The coast had a very different pattern of variation from the simulated sand, in a range with lower values. The water variability in the observed data had a close range to the simulated data. Wood, present only in the observed data, had a range close to that of water.

The mean spectral signatures of each class were very different in the simulated dataset, with the exception of the SWIR2 band (Figure 10, first column). In the observed datasets, the 2019 data, although having a characteristic signature for the coast class, have indistinguishable signatures for water and plastic, which are overlap each other on the chart. The 2021 data show a minimally distinct signature for each class, possibly due to higher percentages of plastic coverage in this set of images (Figure 10, two central columns). The fourth column of the Figure 10 strengthens this hypothesis: it displays mean spectral signatures of water and plastic in the observed data - considering both the 2019 and 2021 sets - but including only those pixels with at least 50% plastic coverage for composition of the mean signature of each class. In this case, the sig natures show a small separability in all bands (with the exception of the SWIR2 band again), differently from what was found in the 2019 dataset.

The mean spectral signatures confirm the optical spectral behavior characteristics described by Themistocleous et al. (2020) and Biermann et al. (2020):

- Clean water is efficient at absorbing light, especially at NIR and SWIR;
- Plastic and wood had a higher reflectance in the Red band and reflectance peaks at NIR;
- Plastics and water demonstrated a higher reflectance in the Blue band, falling in the Green and Red

bands.

However, unlike what was pointed out by the authors, wood presented low reflectance in the SWIR region, as well as plastic, perhaps as an effect of the water absorption, considering that the samples had some degree of submersion in some cases.

The hypothesis proposed by Biermann et al. (2020), that the surface cover classes could be distinguished using the FDI and NDVI indices individually in boxplots and combined in scatterplots was refuted in our results. The water FDI and NDVI values have many outliers and overlap with the values of the other classes. If the outliers are disregarded in the boxplot, there is a separability between water and sand/coast in both datasets, but plastic and wood still have FDI and NDVI values very close to substrate (Figures 11 and 12).

In both datasets, it is possible to differentiate the substrates (water/sand or water/coast) using scatterplots with feature on the Y axis and the identifiers of the pixels on the X axis. However, the plastic and wood pixels appear superimposed on the substrate, without a range of values to discriminate them, contradicting the findings of Themistocleous et al. (2020), that associates plastic pixels with a specific range of PI index values. Here, the index did not demonstrate separability (Figure 13).

The exploratory analysis techniques allowed a better understanding of the characteristics of the datasets, and may provide a knowledge base for other methods in future work.

## FEATURE SELECTION

The feature importances calculated using RF classification are presented in Figure 14. There were no significant differences in feature importance for datasets containing all classes or only water and plastic.

Similar to Basu et al. (2021), different sets of features were created for the classification process. The feature importance scores and correlation values between the features were considered during feature selection. The four sets of features and their selection criteria are presented in Table 3.

**Table 3.** Subsets of features selected for classification.

| ID | Selected features |
|----|-------------------|
| A | All features: full set of spectral bands and full set of indices. |
| B | Full set of spectral bands. |
| C | Full set of indices. |
| D | Features *NIR1*, *SR*, *WRI* and *FDI*. |

Note 1: The full set of spectral bands comprises the bands: *Blue*, *Green*, *Red*, *RedEdge1*, *RedEdge2*, *RedEdge3*, *NIR1*, *NIR2*, *SWIR1* e *SWIR2*.
Note 2: The full set of radiometric indices comprises the indices: *NDWI*, *WRI*, *NDVI*, *AWEI*, *MNDWI*, *SR*, *PI*, *RNDVI* and *FDI*.
Note 3: Set D is composed of the two features with the highest importance score (*NIR1* and *SR*) with the two features least correlated with them (*WRI* and *FDI*). In the correlation analysis for this subset, only features with correlation values between -0.75 and 0.75 were considered, both in the total dataset and in the class-by-class analysis.

The feature importances were calculated for each dataset individually, with the samples divided into training and test subsets. In this scenario, the RF achieved maximum accuracy in classifying the simulated data. However, when trained with simulated data and tested with observed data, the RF only achieved 38% accuracy. Consequently, its use was restricted to feature selection.

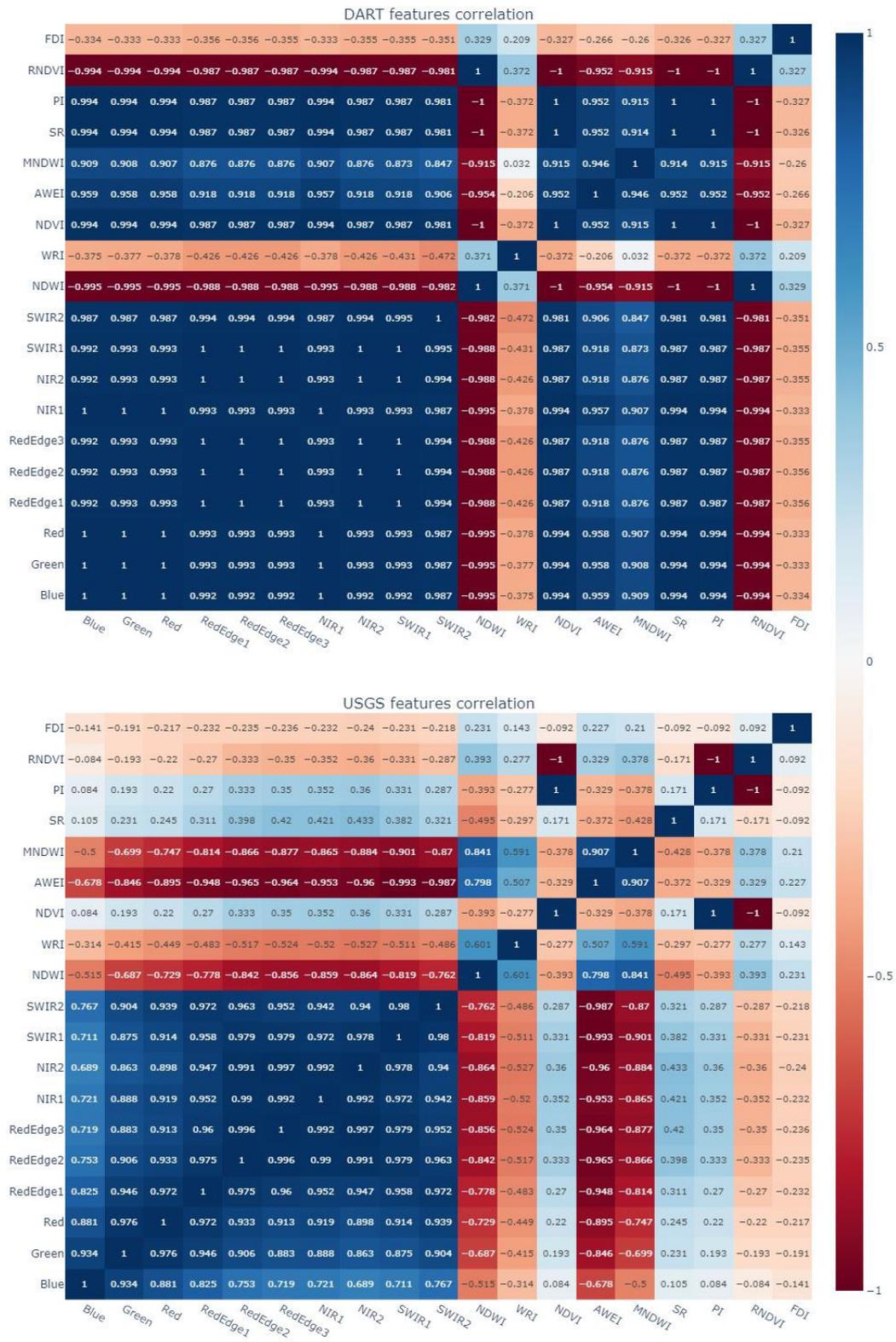

**Figure 6.** Correlation between features in simulated (DART) and observed (USGS) datasets. Positive correlation values are in blue and negative values in red. Darker colors indicate closer proximity to extreme values (-1 for red and 1 for blue).

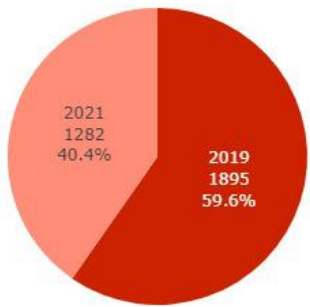

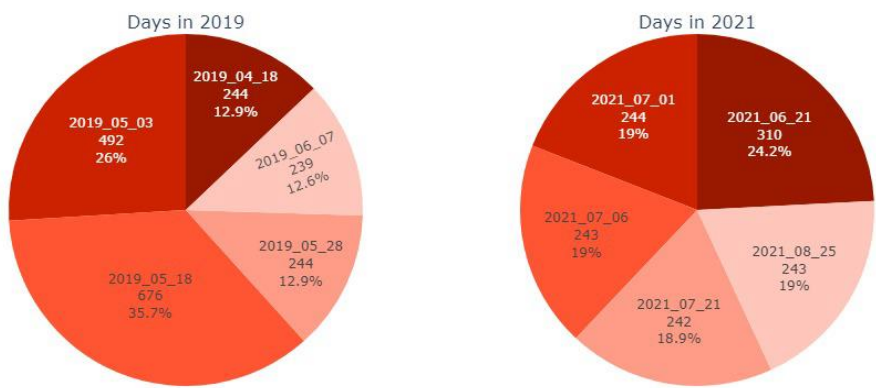

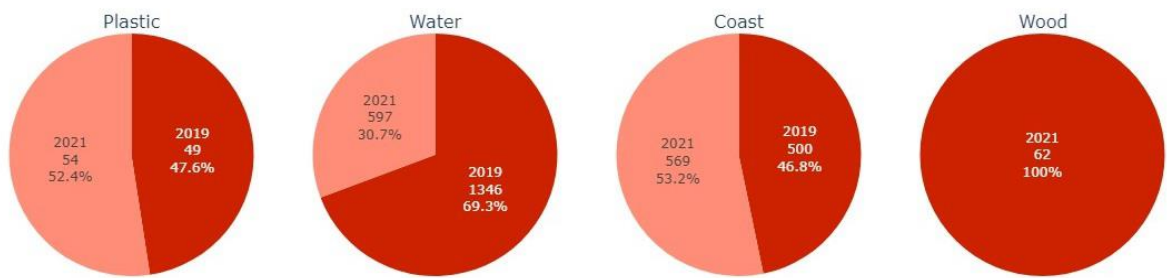

**Figure 7.** Sources and dates of acquisition of the observed dataset (USGS).

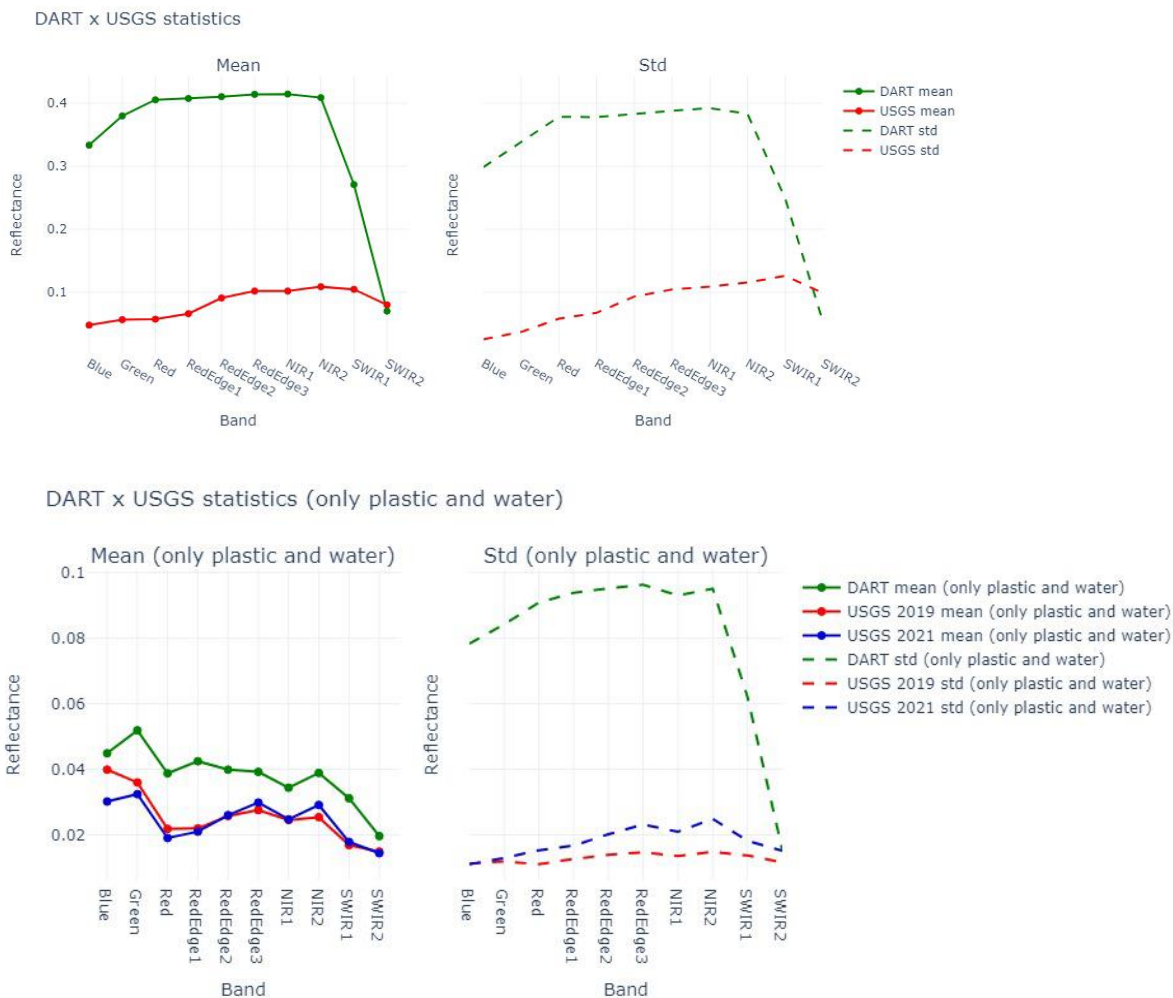

**Figure 8.** Mean and standard deviation in simulated (DART) and observed (USGS) datasets. At top, mean and standard deviation for all classes. At the bottom, mean and standard deviation considering only water and plastic classes.

## UNSUPERVISED CLASSIFICATION

The detailed classification results are presented in the supplementary material. In the unsupervised classification with simulated data, for all tested parameters, more than 95% of samples were grouped in only two clusters mainly made of water and sand respectively, with about half of the pixels of plastic added to each of them. The other clusters had small amounts of water or sand mixed with a few plastic samples. These smaller clusters more often concentrated water subgroups than sand, and when they contained plastic pixels, the samples had at least 60% or 80% of plastic coverage in general. It is a demonstration of the influence of coverage percentages in the clustering process. Pixels with 20% to 60% of plastic coverage were evenly distributed between the two larger groups of water or sand as if they belonged to the same class as the substrate.

Using the observed data, for all tested parameters the algorithm created two clusters adding up to them more than 99% of the total samples, mainly made of water and coast, respectively. The only exception to this tendency was the classification with the feature set B, made of the sensor bands: this configuration concentrated between 50% to 66% of the samples in one cluster, with a predominance of water, and distributed the others samples in percentages somewhat balanced between the other clusters. Figure 15 demonstrates one of the results obtained with feature set B for both data sets.

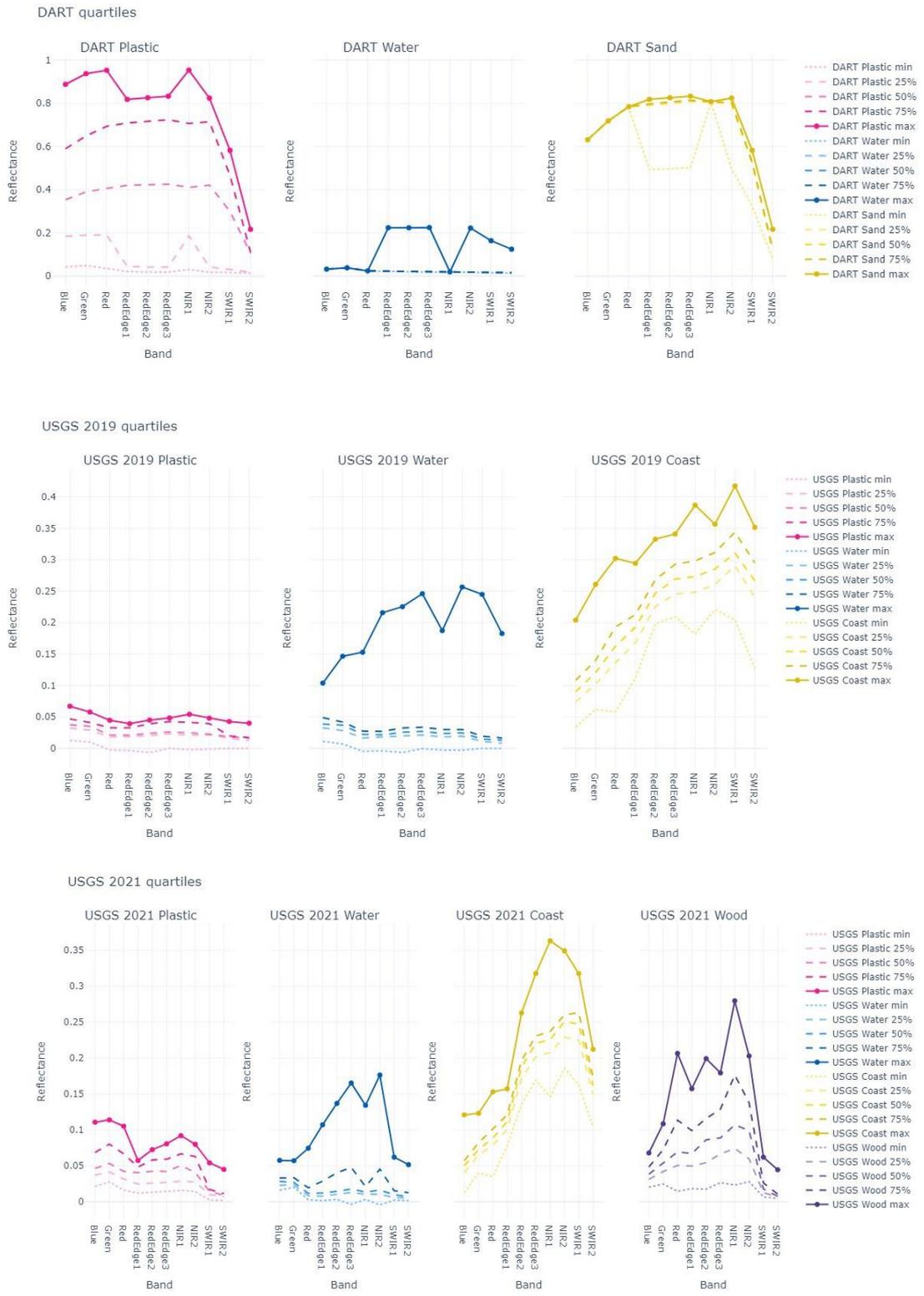

**Figure 9.** Quartiles of the different classes in simulated dataset (DART, top) and in observed dataset (USGS) grouped by year of acquisition (2019 at middle and 2021 at bottom).

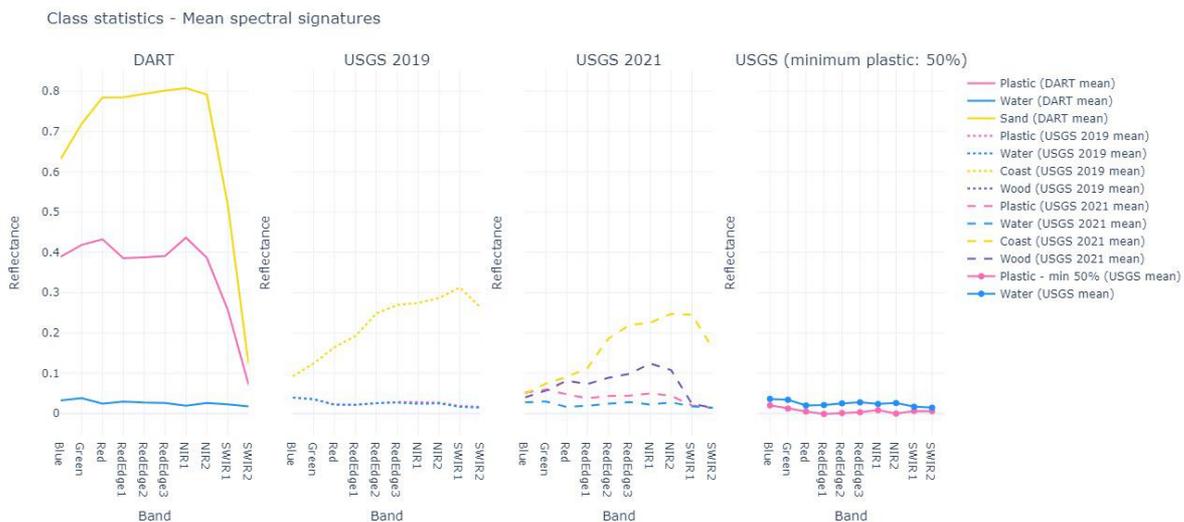

**Figure 10.** Mean spectral signatures of classes grouped by dataset and by coverage percentage.

In the simulated data classification, the water and sand pixels were never placed in the same cluster. In the observed data classification, water and coast were eventually put in the same cluster. The following trends were observed in the simulated data classification: PVC was more frequently associated with sand, while PET was more frequently associated with water; LDPE was generally grouped with water or in exclusive clusters containing few pixels, all with 100% coverage. LDPE was the polymer with the highest tendency to separate from the other elements. Figure 16 demonstrates an example of polymer distribution for feature set B, for both datasets.

These trends confirm the mean signatures of each class from the simulated data displayed at the top of Figure 17: Among the mean signatures of polymers, PET has the lowest mean reflectance (and therefore closest to the signature of water) while PVC has the highest mean reflectance (and therefore closest to the signature of sand). Differences between mean reflectances of polymers are larger when only pixels with 100% coverage are considered.

In the observed data, the totality of pixels with 100% HDPE mesh coverage was always added to clusters with a predominance of water. There were no other detectable patterns of association between elements or percentage coverage. The water groups tended to include wood and plastic pixels, and the coast groups tended to isolate, grouping with other classes less frequently.

The distribution of image acquisition dates among the clusters was also analyzed. The only identifiable influence of acquisition dates on cluster formation was the separation between images from 2019 and 2021 in some cases. Any of the classification configurations had separated different dates of the same year into different clusters, so there is no evidence of impact caused by differences in target conditions at different dates of the two PLPs, such as different rates of submersion and biofouling pointed out by Papageorgiou et al. (2022) in the 2021 images.

Responding to the questions that guided the analysis:
- The number of clusters did not affect the classification for any feature set.
- The association patterns between pixels were influenced by type of coverage (in pixels containing only substrate) and by type of polymer and percentage of coverage (in pixels containing plastic or wood).
- Different feature sets and number of clusters did not generate notable differences in class separability.

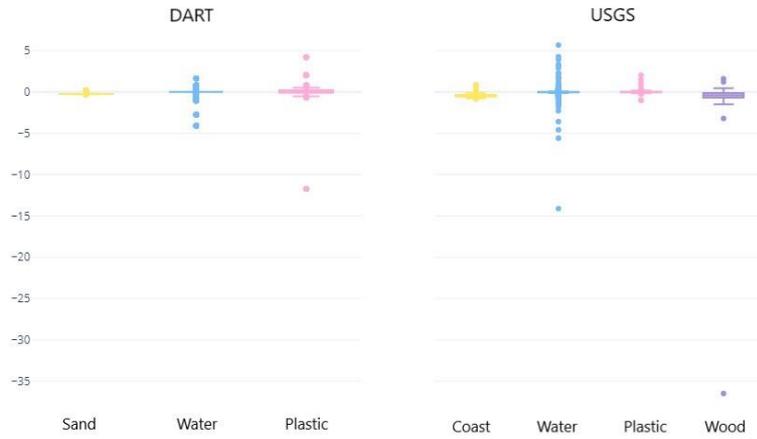

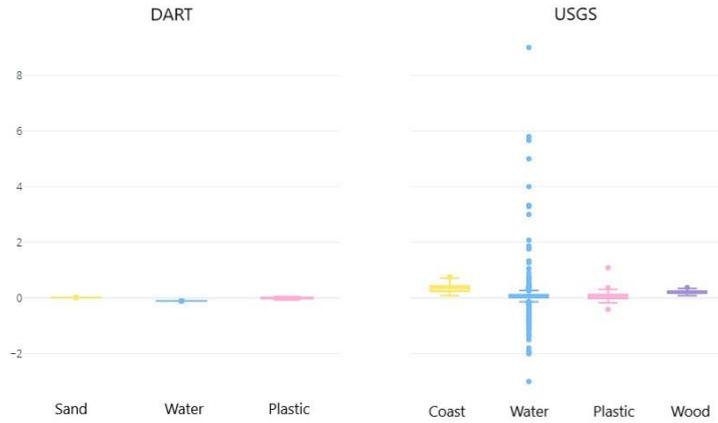

**Figure 11.** FDI and NDVI values grouped by class in the simulated (DART) and observed (USGS) datasets.

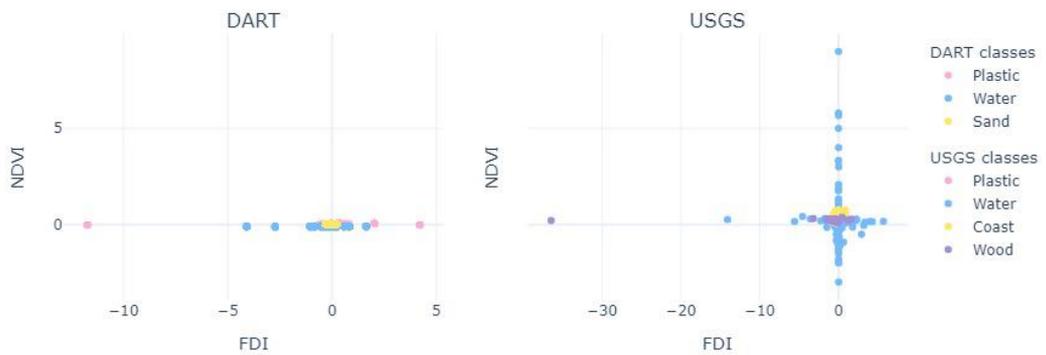

**Figure 12.** Scatterplot of FDI and NDVI values in the simulated (DART) and observed (USGS) datasets.

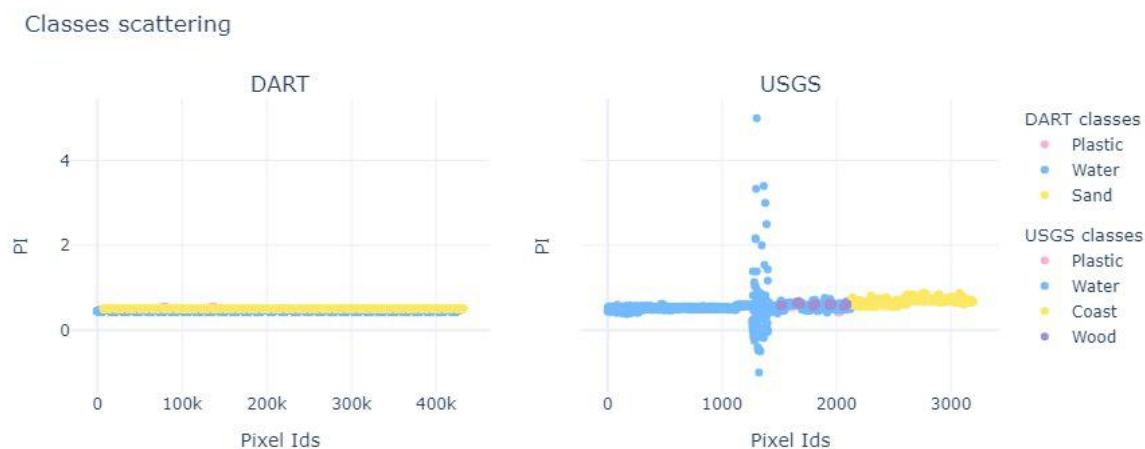

**Figure 13.** Scatterplot of PI values grouped by class in the simulated (DART) and observed (USGS) datasets.

## DISCUSSION

The differences between the simulated and observed datasets in the exploratory analysis highlighted the statistical differences between the sets. The sets belong to different statistical distributions, violating one of the basic assumptions of machine learning and potentially limiting its application in this context.

Considering that the simulations had fewer elements and greater spectral purity, the grouping of different substrates in the same cluster occurring only in the observed data (in this case, sand and coast) is an expected result. It is a positive point that this clustering was not frequent in the observed data because it shows the algorithm can differentiate the surface cover classes, even in an unfavorable scenario.

The differentiation between plastic or wood targets and the substrate tends to occur only from higher percentages of pixel coverage, which shows a limitation of the classifier concerning the spatial resolution of the sensor. At the same time, the plastic or wood pixels with higher coverage percentages separation from the substrate in different clusters show the classifier's capacity for detection, despite the limitations imposed by spatial resolution.

The results are not definitive about the applicability of RTMs. Partial discrimination of plastic and wood elements in pixels with high coverage percentages, both in simulated and observed data, can be considered a favorable indication for the use of RTMs, considering that there was detectability in sets that belonged to statistical distributions different and therefore limited the potential of machine learning. However, further study is needed for validation. We recommend that, if possible, new applications use simulations created with more diverse spectral databases, generating synthetic scenes closer to the diversity of observed scenes, and consequently creating more favorable conditions for the classifier.

The unique behavior in the classification using feature set B as input, which contained only the sensor bands, demonstrates that using or not using radiometric indices can be more decisive for the result than using one or another given index. The results also contradict the findings of other studies using radiometric indices to detect plastic pollution (Biermann et al. 2020, Themistocleous et al. 2020). Considering that the use of indices led to a higher concentration of samples, its use may not be indicated for classification of marine plastic, but further studies are needed for validation.

The study of spectral behavior carried out through exploratory analysis and unsupervised classification allowed the determination of spectral characteristics of elements and trends of association and differentiation between elements. These findings may guide further research in remote sensing applied to marine plastic pollution detection tools.

At the same time, the applied methodology is strongly dependent on the data. New spectral databases were made available during the development of this work (Knaeps et al. 2020), and may be used in future work to create simulations closer to the observed conditions, potentially generating even better results.

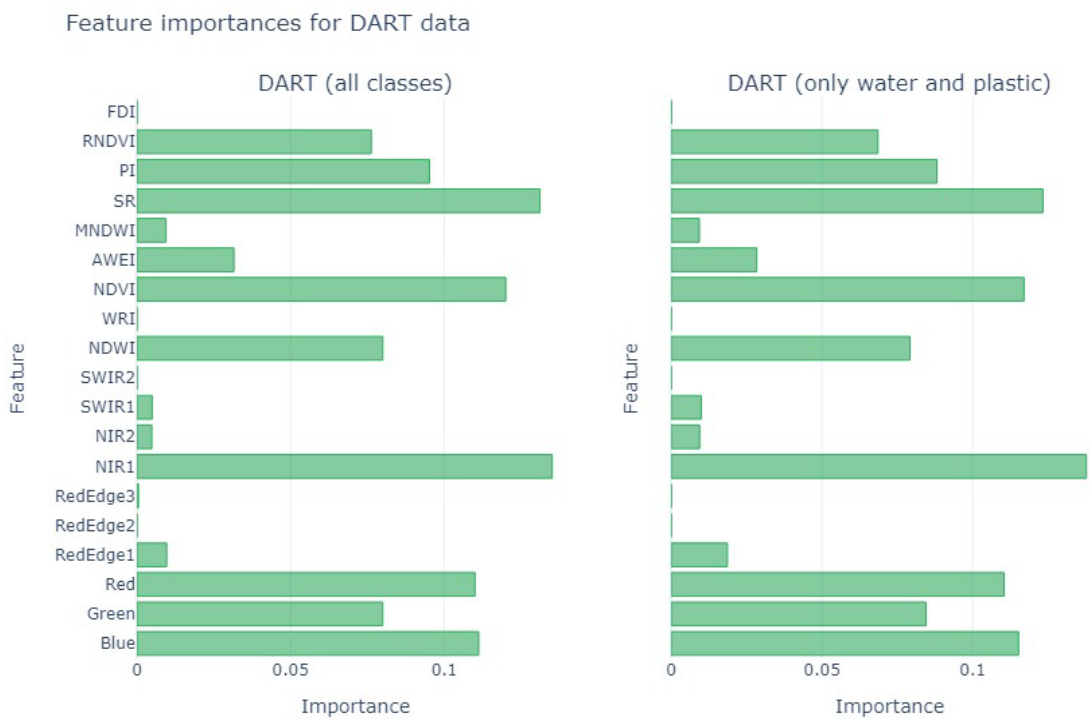

**Figure 14.** Feature importances for the simulated (DART) dataset calculated by the RF algorithm.

## CONCLUSION

The methodology developed met the objectives of the work. The exploratory analysis contributed to the objective of studying the spectral behavior of marine pollution by plastics, demonstrating in detail the spectral characteristics of the elements present in the scenes. The unsupervised classification complemented these results, showing similarities, contrasts, and patterns of association.

It was possible to carry out a partial detection of the target elements, but depending on the spectral purity and the percentage of coverage of the pixels. It is needed in future work to seek strategies to deal with cases of spectral mixing, noise, and the need for detection at the subpixel level, imposed by the limitation of sensor resolutions. Possible approaches include spectral separation models, image super-resolution techniques, specific methodologies for treating data imbalance, and creating simulations with a greater variety of parameters.

Improvements in the available datasets will be decisive for machine learning accuracy. New spectral databases are needed to create new simulations, with greater variability of elements and conditions (i.e. Knaeps et al. 2020), allowing the inclusion of factors such as submersion and bioencrustation in synthetic scenes. It is also important to create new highquality observed datasets with in situ validation, for more robust analyzes of the factors that influence the detectability of marine pollution.

# DART - Features set B

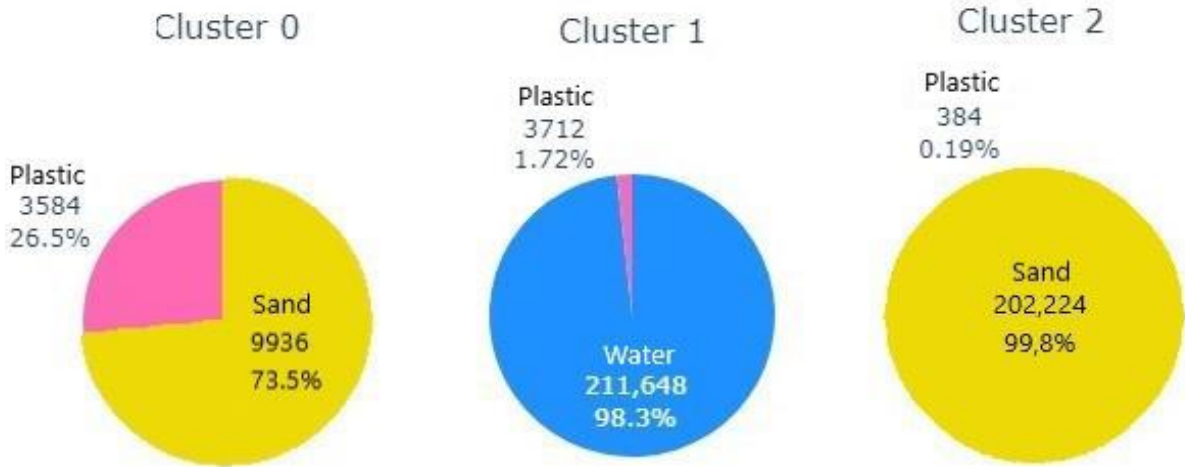

# USGS - Features set B

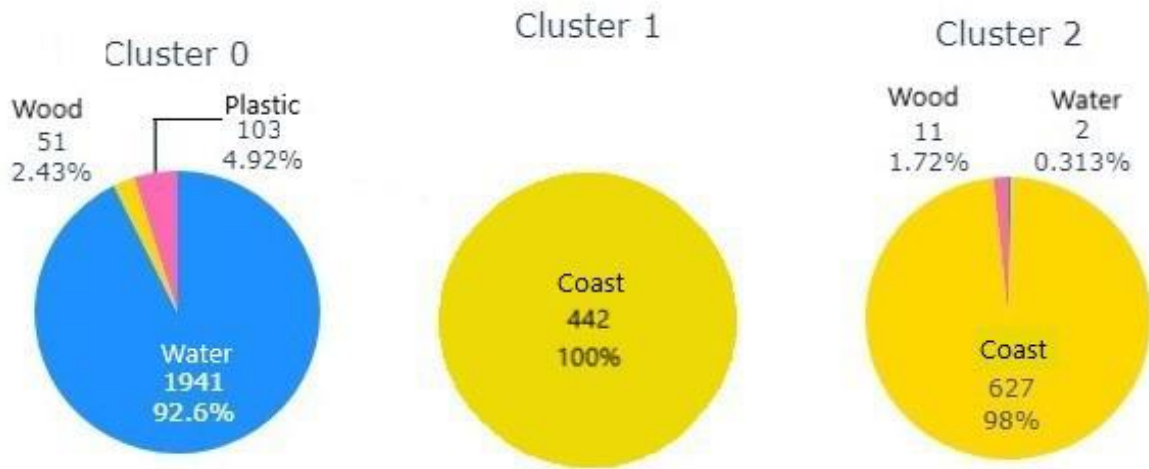

**Figure 15.** K-means algorithm clusters for feature set B.
Note: Clustering result for feature set B (spectral bands) using k=3 for simulated (DART) and observed (USGS) datasets. The percentages of samples per class refer to the proportions of the cluster, not the total dataset (to see the proportions of each cluster in relation to the total dataset, see supplementary material).

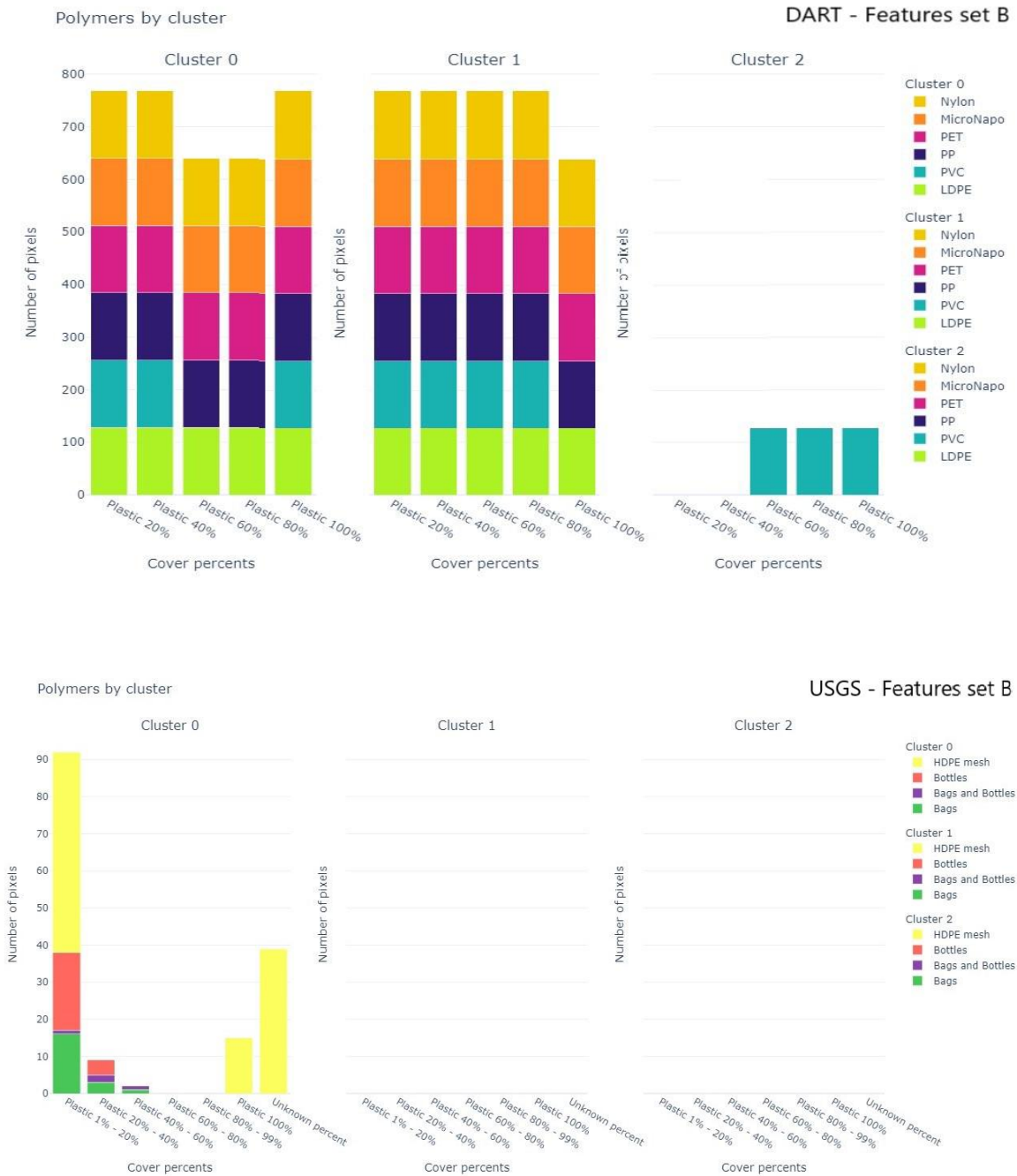

**Figure 16.** Polymers distribution in K-means clusters for feature set B.
Note: Clustering result for feature set B (spectral bands) using k=3 for simulated (DART, top) and observed (USGS, bottom) datasets.

## ACKNOWLEDGMENTS

The authors thank Universidade Federal do Rio Grande do Sul, for providing the environment for the research development. The authors also thank Toulouse III University for licensing the DART computational tool, as well as the European Space Agency and the University of the Aegean for making publicly available the observed images used in this work.

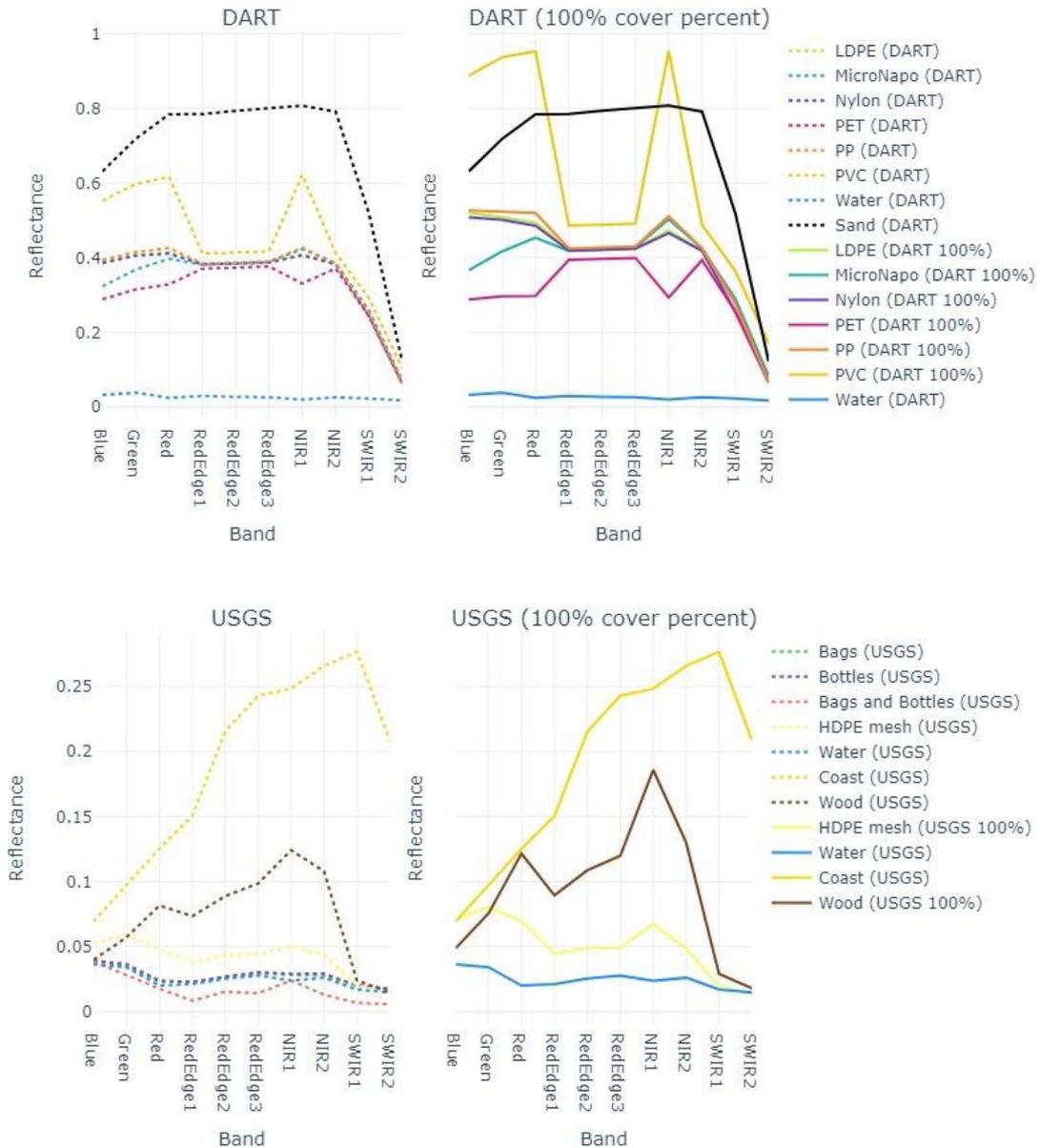

**Figure 17.** Mean spectral signatures by class for simulated (DART, top) and observed (USGS, bottom) datasets.

## AVAILABILITY OF SOURCE CODE AND REQUIREMENTS

- Project name: Plastic Map
- Project home page: https://github.com/ufrgs-rescue/plastic_map
- Operating system: Platform independent
- Programming language: Python
- Other requirements: Python libraries datetime, numpy, os, pandas, plotly, rasterio, scipy and sklearn
License: GNU GPL

## AVAILABILITY OF SUPPORTING DATA AND MATERIALS

The data supporting the results of this article are available in the Plastic Map repository [https://zenodo.org/record/8083619].

## DECLARATIONS

### LIST OF ABBREVIATIONS

List of abbreviations in alphabetical order:
- AWEI: Automated Water Extraction Index
- DART: Discrete Anisotropic Radiative Transfer
- FDI: Floating Debris Index
- GPS: Global Positioning System
- HDPE: High Density Poly Ethylene
- LDPE: Low-density polyethylene
- MNDWI: Modified Normalization Difference Water Index
- MRSG: Marine Remote Sensing Group
- MSI: Multispectral Instrument
- NDVI: Normalized Difference Vegetation Index
- NDWI: Normalized Difference Water Index
- NIR: Near Infrared region of the electromagnetic spectrum
- PA 6 and PA 6.6: Polyamide 6 and 6.6 or Nylon
- PET: Polyethylene terephthalate
- PI: Plastic Index
- PLP: Plastic Litter Project
- PP: Polypropylene
- PVC: Polyvinyl chloride
- RE: Red Edge region of the electromagnetic spectrum
- RF: Random Forest Algorithm
- RNDVI: Reversed Normalized Difference Vegetation Index
- RTM: Radiative Transfer Model
- SNAP: Sentinel Application Platform
- SR: Simple Ratio
- SWIR: Shortwave Infrared region of the electromagnetic spectrum
- VIS: Visible region of the electromagnetic spectrum
- UAV: Unmanned Aerial Vehicle
- WRI: Water Ratio Index
- μ-NAPO: Mean signature made of several microplastics collected in the Pacific Ocean

### COMPETING INTERESTS

The authors declares that they have no competing interests.

### AUTHOR'S CONTRIBUTIONS

The author contributions according to the Contributor Roles Taxonomy CASRAI CRediT:

• Bianca Matos de Barros: Conceptualization, Data Curation, Formal Analysis, Investigation, Methodology, Resources, Software, Visualization and Writing – Original Draft.

• Douglas Galimberti Barbosa: Data Curation, Investigation, Resources and Writing – Review & Editing.

• Cristiano Lima Hackmann: Conceptualization, Methodology, Project Administration, Supervision and Writing – Review & Editing.